\documentclass[runningheads]{llncs}

\usepackage{booktabs}
\usepackage{multirow}
\usepackage{color}
\usepackage[table]{xcolor}
\usepackage{tikz}
\usepackage{pgfplots}
\usetikzlibrary{plotmarks}
 
\newcommand\blue[1]{{\color{blue}#1}} 
\newcommand\green[1]{{\color{green}#1}} 

\usetikzlibrary{arrows.meta}

\usepackage{algorithm}
\usepackage{algpseudocode}

\usepackage{amsfonts,amsmath,amssymb}
\usepackage[utf8]{inputenc}
\usepackage{graphicx}
\usepackage{verbatim} 

\newcommand*{\field}[1]{\mathbb{#1}}%

\usepackage{multirow}
\usepackage{booktabs}
\usepackage{url}
\usepackage{subfig}
\usepackage{dblfloatfix} 
\usepackage{cite}

\usepackage{graphicx}
\usepackage{setspace}
\usepackage{multicol}

\usepackage{blindtext}

\usepackage{tikz}
\usepackage{url}
\urldef{\mailsa}\path|{giovanni.colacurcio, sergion}@fing.edu.uy|
\urldef{\mailsb}\path|flv@lcc.uma.es|
\urldef{\mailsc}\path|toutouh@mit.edu|
\urldef{\mailsd}\path|diego.rossit@uns.edu.ar|

\pgfplotsset{compat=1.15}

\begin{document}

\title{Multiobjective household energy planning \\using evolutionary algorithms}
\author{
Giovanni Colacurcio\inst{1} \and Sergio Nesmachnow\inst{1} \and \\Jamal Toutouh\inst{2} \and Francisco Luna\inst{3} \and Diego Rossit\inst{4}}
\authorrunning{G. Colacurcio, S. Nesmachnow, J. Toutouh, F. Luna, D. Rossit} 
\tocauthor{Giovanni Colacurcio, Sergio Nesmachnow, J. Toutouh, Francisco Luna, Diego Rossit}
\institute{Universidad de la Rep\'{u}blica, Uruguay,\\
\mailsa \vspace{2mm}
\and Massachusetts Institute of Technology, %CSAIL, MA, 
USA \\
\mailsc  \vspace{2mm}
\and Universidad de Málaga, Spain\\
\mailsb  \vspace{2mm}
\and
Universidad Nacional del Sur, Argentina\\
\mailsd
}

\maketitle

\begin{abstract}
This article presents the advances in the design and implementation of a recommendation system for planning the use of household appliances, focused on improving energy efficiency from the point of view of both energy companies and end-users. The system proposes using historical information and data from sensors to define instances of the planning problem considering user preferences, which 
in turn are proposed to be 
solved using a multiobjective evolutionary approach, in order to minimize energy consumption and maximize quality of service offered to users. Promising results are reported on realistic instances of the problem, compared with situations where no intelligent energy planning are used (i.e., `\textit {Bussiness as Usual}' model) and also with a greedy algorithm developed in the framework of the reference project. The proposed evolutionary approach was able to improve up to 29.0\% in energy utilization and up to 65.3\% in user preferences over the reference methods.
\end{abstract}
%\keywords{smart cities \and timetabling \and evolutionary computation.}

\section{Introduction}
%[Contexto]
Energy management is a crucial issue in nowadays societies. Many strategies have been proposed to guarantee an increased access to the energy resources at affordable costs for citizens, while ensuring the preservation of natural resources and the protection of the environment~\cite{Turner2007}.

For 
%the implementation of 
implementing
effective energy management policies, innovative technologies must be integrated in easy-to-use and efficient systems,
%Energy management systems need to 
which must 
include specific features to be useful for both energy companies and citizens. In order to be applied by energy companies, management systems must include capabilities for performing realistic simulations,
%and also for 
controlling, and planning the electricity market. From the point of view of citizens, system must provide easy-to-use applications to monitor and manage the energy consumption at household level. 

%The capabilities of monitoring, controlling, and managing the energy consumption and generation are very important to provide good quality of service (QoS) and user experience; especially, when considering the emphasis on citizen engagement, environment protection, and economic considerations, under the novel paradigm of smart cities~\cite{Deakin2011}.

The capabilities of monitoring, controlling, and managing the energy consumption and generation are very important to provide good quality of service (QoS) and user experience, especially when considering the emphasis on citizen engagement, environment protection, and economic considerations, under the novel paradigm of smart cities~\cite{Deakin2011}.

%[Motivación]
Residential buildings significantly contribute to the total energy used in the world. According to statistics from the USA Energy Information Administration, the average household in the USA and Canada uses about 12,000 kWh of electricity. In Europe the figure is less than 10,000 KWh, but also significant. Some household appliances make the biggest contributions to consumption, including heaters and air conditioning (40--45\%), electronic and kitchen appliances ($\sim$30\%), water heaters (15\%--20\%), and cooling (5\%--10\%)~\cite{EIA2018}. Energy utilization patterns are not too different in developing countries, where the impact of energy consumption is also very important.

The related literature indicates that systems based on planning the use of 
%\textit{deferrable loads} or 
\textit{deferrable appliances} allow improving energy efficiency at domestic level~\cite{Soares2013, Soares2014}.
Deferrable appliances are those whose demand for energy can be postponed or interrupted (such as dishwashers, washing machines, etc.), causing a negligible impact on the QoS provided to users. This is an important approach that takes into account the different prices of electricity and the availability of (non-storable) energy from renewable sources.

%[Propuesta] 
In this line of work, this work presents the application of evolutionary algorithms (EAs) to solve the problem of planning the use of household appliances considering user preferences. A multiobjective approach is considered, aiming at simultaneously maximizing user satisfaction (evaluated in terms of the QoS offered according to the specified preferences) and minimizing the total energy consumed, which is directly related to the total cost of the electricity bill for the user.
%
%[Contribuciones]
The main results indicate that the proposed approach is able to find appropriate plannings that improve over situations where no intelligent energy planning are used (i.e., `\textit {Bussiness as Usual}' model) and also with a greedy algorithm previously proposed in the framework of the reference project.

The main contributions of this research are: 
\textit{i)} formulating a novel multiobjective household energy planning problem accounting for user satisfaction and energy consumption; 
\textit{ii)} devising a specific EA to address the problem, using a linear aggregation approach for the 
%problem 
objectives; and 
\textit{iii)} evaluating the proposed evolutionary approach over realistic scenarios, built by using real data of household energy consumption from well-known repositories.

The research reported in this article is developed in the context of the Cloud Computing for Smart Energy Management (CC-SEM) project~\cite{Lujan2018,Lujan2019}, which proposes building an integrated platform for smart monitoring, controlling, and planning energy consumption and generation in urban scenarios. 
The project integrates cutting-edge technologies (Big Data analysis, computational intelligence, Internet of Things, High Performance Computing and Cloud Computing) and specific hardware for energy monitoring/controlling built within the project. 

The article is organized as follows. Section~\ref{Sec:Prob} presents the formulation of the multiobjective household energy consumption planning and a review of related works. The proposed evolutionary approach for household energy planning is described in Section~\ref{Sec:Alg}. The experimental analysis is reported in Section~\ref{Sec:Exp}. Finally, Section~\ref{Sec:Conc} presents the conclusions and the main lines of future work.

\section{The household energy planning problem}
\label{Sec:Prob}

This section introduces the household energy planning problem, the multiobjective formulation addressed in this article, and a review of related works.
%about household energy planning.

\subsection{General considerations}
The goal of the study is to 
%apply computational intelligence methods 
develop a system
to help end-users to take appropriate decisions concerning the use of household appliances 
%to be managed 
in a given planning period (e.g., daily, weekly, etc.). 
The problem consists in scheduling the use of different household appliances to minimize the electric bill, taking into account the end-user preferences, electricity prices, and the available contracted power.

The planning period is divided in slots 
%to take into account 
considering
the user preferences. For every slot, each user can indicate a value that represents the priority of using a certain appliance in that time. Higher values of priority represent a higher desire 
%of the user 
of using the appliance. In case that users do not indicate 
%explicitly 
their preferences, machine learning 
%techniques are 
is
applied to infer preferences from the analysis of historical utilization
data.
%of domestic appliances. 
Classification methods can also be applied 
%in order 
to characterize the 
%power consumption of each house, 
household power consumption,
regarding neighboring houses and 
%also using 
socio-economical data, 
%in a similar way that characterization of 
such as for
other public services~\cite{Nesmachnow2017}.

The problem formulation assumes that the energy cost is known for each time interval. These values are publicly available from the 
energy companies, for example from the National Electricity Company (UTE) in Uruguay. %(\url{https://portal.ute.com.uy}).
Also, the maximum contracted power for each user is known, from the contract details provided by the energy company.
The contracted maximum power can only be surpassed by a small amount (up to 30\%) in a short period of time, without causing a short circuit. Schedules that includes such a surplus are penalized.

%Our approach base on create a multiobjective genetic algorithm that takes into account all the information that we discussed before and return the allocation of loads for the given planning period. In addition we studied mono-objective genetic algorithms that considers prices, contracted electric power and QoS.

%Para la revista
%Real life case studies have been investigated to analyze the energy consumption characteristics for various urban levels (houses, buildings, neighborhoods, etc.)

\subsection{Problem formulation}

The multiobjective version of the household energy planning problem addressed in this article considers the following elements:

\renewcommand{\labelitemi}{$\bullet$}
\begin{itemize}
  \item a set of users  $U = \left(u_{1} \ldots u_{N} \right)$, each user represents a 
  %housing unit (house) 
  house
  in a city;
  \item a set of time slots $T = \left(t_{1} \ldots t_{M} \right)$ in the planning period;
  \item a set of domestic appliances 
  %(generically named \textit{loads}) 
  $L = \left(l_{1} \ldots l_{K} \right)$ for each user;
  \item a function $E:U\rightarrow \field{N}$, where $E(u_i)$ indicates the maximum electric power contracted by user $u_i$;
  \item a penalty term $\rho$ applied to those users that surpass the maximum electric power contracted;
  \item a function $D:L \times U \rightarrow \field{N}$, where $D(l_k,u_i)$ indicates the average time of utilization of appliance $l_k$ for user $u_i$;
  \item a function $C:T\rightarrow \field{N}$, where $C(t_j)$ indicates the utilization cost (per kW) of the energy in time $t_j$;
  \item a function $P:L\rightarrow \field{N}$, where $P(l_k)$ indicates the power (in kWh) consumed by appliance $l_k$;
  \item and a function ${UP}:U \times L \times T \rightarrow \field{N}$, where $UP(u_i,l_k,t_j)$ indicates the preference of user $u_i$ to use the appliance $l_k$ at time $t_j$;
\end{itemize}

Lets consider the binary variable $x^{i}_{kj}$, that indicates if appliance $l_k$ of user $u_i$ is turn on at time $t_j$;
and function
$y(x^{i}_{kj})$ that indicate the time period in which appliance $l_k$ of user $u_i$ is turned on continuously (without intermediate turn off) from time $t_j$:
%(Eq.~\ref{eq:per})
%
%\begin{eqnarray}
$
y(x^{i}_{kj}) = m - j \textrm{ with } m = \max{r / \forall h \in (j,r) \textrm{ } x^{i}_{kh} = 1 }
$
%\label{eq:per}
%\end{eqnarray}

The problem proposes finding a planning function $X = \{x^{i}_{kj}\}$ for the use of each household appliance that simultaneously maximizes the user satisfaction defined in Eq.~\ref{eq:pref} (given the users' preference functions)
%, minimize the cost of the electric bill (Eq.~\ref{eq:cost}) 
and minimize the total energy consumed (see Eq.~\ref{eq:cons}).

%$f(X)$ \red{falta x} que retorna las preferencias de usuario para los aparatos que están actualmente encendidos \[ 
\begin{eqnarray}
f(X) = \sum_{i=1}^{N}\sum_{j=M}^{T}\sum_{k=1}^{K} UP(u_i,l_k,t_j) \times \delta^i_{kj} 
\label{eq:pref} \\
\nonumber
\textrm{with} \enskip \delta^i_{kj} = 
\begin{cases}
	\textrm{1} \enskip \textrm{if} \enskip y(x^{i}_{kj}) \geq D(l_k,u_i)\\
    0 \enskip \textrm{otherwise}
\end{cases} 
%\tag{1}
\end{eqnarray}

%A function $h$ that indicate the total cost \[
%\begin{eqnarray}
%h(X) = \sum_{i=1}^{N}\sum_{j=M}^{T}\sum_{k=1}^{K} x^i_{kj} \times P(l_k) \times C(t_k) 
%\tag{3}
%\label{eq:cost}
%\end{eqnarray}

%A function $g$ that indicates the total Power Consumed \[
\begin{eqnarray}
\enskip g(X) = \sum_{i=1}^{N}\sum_{j=M}^{T}\sum_{k=1}^{K} x^i_{kj} \times C(t_j) \times P(l_k) + \rho \times \psi^i_{kj}
\label{eq:cons}
\\
\nonumber
\enskip \textrm{with} \enskip \psi^i_{kj} = 
\begin{cases}
	P(l_k) - E(u_i) \enskip \textrm{if} \enskip P(l_k) - E(u_i) \geq 0\\
    0 \quad \textrm{otherwise}
\end{cases} 
%\tag{2}
\end{eqnarray}

Two scenarios are defined for defining the penalty model used for those situations in which the household consumption exceeds the maximum power contracted. The first scenario (\textit{soft penalty}) is when the user exceeds the maximum power contracted for less than 30\% of it. This is the maximum value of energy consumption that can exist without a short circuit occurring. In that case, the solution is penalized by a 30\% of $\rho$. The second scenario (\textit{hard penalty}) is when the user exceeds the maximum power contracted in a value greater than or equal to 30\%. Therefore, that plannings are penalized entirely by the penalty term $\rho$.

Function $UP$ considers the energy consumption measurements of electrical devices reported by Kolter and Johnson~\cite{Kolter2011}. For each minute of the day, in the period of a month, the user preference is defined considering how many times each appliance was turned on for each appliance at that minute.

Function $D$ uses consumption values of user appliances from a representative day. The duration of using for the appliance was studied, defined as the number of consecutive minutes in which it remained powered on~\cite{Chavat2019}.

%%%%%%%%%%%%%%%%%%%%%%%%%%5
\subsection{Related works}
\label{Sec:RW}

The analysis of the related literature allows identifying several 
%recent 
%proposals 
%on the design and application 
%of 
hardware- and software-based methods for household energy consumption characterization and planning. The main related works are reviewed next.

The main line of work related to the proposed research has been developed by Soares et al., who studied the 
%problem of the management of domestic demand 
household electricity demands
and categorized a set of 
%household 
appliances,
%in common use, 
according to their use and management strategies that can be applied to them~\cite{Soares2014b}. An initial work~\cite{Soares2013} introduced a model based on integer non-linear programming for %the problem of 
energy utilization planning, with the aim of reducing cost. The authors applied an 
%evolutionary technique 
EA
to minimize the cost of invoice and violations to the maximum contracted power. The EA allowed to reduce up to 40\% the energy cost for the users with respect to a reference scenario without demand management. Later, the authors 
%also presented MOEA to minimize the cost and maximize 
proposed minimizing cost and maximizing 
user satisfaction~\cite{Soares2014}, which is the main motivation for the work proposed in our research. 
%The main
Results showed that the cost reduction was 
%between 22\% and 24\%, depending on the levels of satisfaction. 
22--24\%.
However, 
%the monoobjective approach does not calculate compromise solutions, so that
no trade-off solutions were computed, so
different users with equal contracted power and equal preferences should adapt to the same % load 
planning. 
%In the multiobjective approach, the penalties of user preference and the penalty of reaching the maximum energy contracted are weighed in a single objective function, which reduces the information provided by the results of the algorithm. 
Additionally, 
%none of the studies 
no studies
were carried out in different urban levels (buildings, neighborhoods, etc.) or used real data. 
%
%QUEDA PARA EXTENSIÓN ! elaborar una frase con esto
%Experiments were based on synthetic data and no real system was proposed to implement demand management in households.

Our previous work~\cite{Orsi2017} presented a hardware and software platform for intelligent monitoring and planning of energy consumption in 
%urban facilities. 
homes.
The proposed system integrates a hardware controller for energy efficiency, a communication protocol to improve data transmission, and a software module for planning and managing household devices. The proposed solution was implemented applying the Internet of Things (IoT) paradigm, allowing the integration of computational intelligence techniques. A greedy algorithm was proposed for planning, considering user preferences and a maximum allowed power consumption. Results showed that it is possible to reduce the energy consumption of a water heater to 38.9\% and that two water heaters and an air conditioner can be optimized simultaneously without reducing QoS. These results suggest that the proposed approach is useful for energy consumption planning in homes.

Bilil et al. proposed a characterization of household appliances and a dynamic planning method for collaborative microgrids~\cite{bilil2016dynamic}. Two multiobjective optimization problems were studied, accounting for the activation and power profiles of appliances. 
A simulation procedure was applied to generate the instances of these problems and NSGA-II was used to solve them. 
The instances consisting in 40 microgrids that inclulde a flexible deferrable appliance, such as a water heater, and a non-flexible one (i.e., dishwasher).  
For the experiments, a residential load curve based on U.S. user profiles was used. 
The results showed that the load curve can indeed become very flat by applying the proposed bi-level multiobjective optimization scheduling approach. 
%The results verify that the load curve can indeed become very flat by applying the proposed scheduling approach.
%https://www.ncbi.nlm.nih.gov/pmc/articles/PMC5562235/
%doi: 10.1109/TPWRS.2016.2613479

The analysis of the related works indicates that there is room to contribute with solutions focused on the development of systems to implement the management of domestic demand through the integration of IoT technologies and computational intelligence algorithms.

\section{The proposed EA for household energy planning}
\label{Sec:Alg}

This section describes the proposed EA to solve the household energy planning problem.

% PARA LA REVISTA
%\subsection{Multiobjective evolutionary algorithms and NSGA-II}
\label{Sec:MOEA}
\subsection{Evolutionary algorithms}
\label{Sec:EA}

EAs are stochastic techniques that emulate 
%the process of 
natural evolution 
%of species 
to solve optimization, search, and learning problems.
%~\cite{BFM97}.
%They have been successfully applied to solve optimization problems in the last 30 years. 
They are 
%especially 
useful for solving complex real-world problems in multiple application areas~\cite{Nesmachnow2014}. 

An EA is an iterative technique (each iteration is called \textit{generation}).
In each generation, probabilistic operators are applied on a set of individuals (the \textit{population}).
The initial population is generated by applying a random procedure or using a specific heuristic for the problem to be solved.
Each individual encodes a tentative solution to the problem and has a {\em fitness} value that determines its suitability to solve the problem. The goal of the EA is to improve the fitness of individuals in the population.
In order to achieve this objective, \textit{evolutionary operators} are applied iteratively, such as the \textit{recombination} of parts of two individuals and the random \textit{mutation} of an individual's coding.
These operators are applied to individuals selected according to their fitness, thus guiding the EA toward tentative solutions of higher quality that replace old individuals.
The stop criterion usually involves a fixed number of generations, a quality level on the fitness of the best individual, or detecting convergence.
The EA returns the best solution found in the iterative process, taking into account the fitness function considered for the problem.
Algorithm~\ref{Alg:EA} presents the generic schema of an EA with a population \textit{P}.

\begin{algorithm}[!h]
\small
\caption{Schema of an evolutionary algorithm.}
\label{Alg:EA}
\begin{algorithmic} [1]
\State {\bf initialize}($P$(0)) 
\State $t$$\leftarrow$0 \Comment{\blue{generation counter}}
\While{not stop criterion}
\State {\bf evaluate}($P$($t$)) \Comment{\blue{evolutionary cycle}}
\State parents$\leftarrow${\bf selection}($P$($t$))
\State children$\leftarrow${\bf variation operators}(parents)
\State newpop$\leftarrow${\bf replacement}(children,$P$($t$))
\State$t$++
\State$P$($t$)$\leftarrow$newpop
\EndWhile
\State {\bf return} best individual found \Comment{\blue{best fitness value}}
\end{algorithmic}
\end{algorithm}

\if 0
The stop criterion usually involves a fixed number of generations, a quality level on the fitness of the best individual, or detecting convergence.
%Additionally, the policy of \emph{selection} of individuals to participate in the recombination and the policy of \emph{replacement} to determine which new individuals are inserted into the population in each new generation must be specified. 
The \emph{selection} policy determines the individuals participating in the evolution and the  \emph{replacement} policy determines the new individuals inserted into the population in each new generation. 
Finally, the EA returns the best solution found in the iterative process, taking into account the fitness function considered for the problem.
\fi

In this article, a traditional EA using a linear aggregation approach is used to solve household energy planning problem.
Although the aggregation approach is often outperformed by Pareto-based methods for multiobjective optimization, it is a common approach in the literature, because of two main advantages: i) it is suitable for multiobjective optimization problems with a convex Pareto front, and ii) it is computationally efficient, so it is recommended when the times available to perform the planning is short~\cite{Coello2002}.

% PARA LA REVISTA
%\subsection{The proposed NSGA-II algorithm for household appliances planning}
\subsection{The proposed EA for household appliances planning}

\vspace{-1mm}
The main features of the proposed EA for household appliances planning are described next.

\vspace{2mm}
\textit{Solution encoding.}
A problem-specific encoding is used to represent solutions. The proposed encoding considers for each user a vector $X = (x_{0}, x_{1},\ldots,x_{T})$, where $T$ is the total number of timesteps (i.e., minutes) in the planning period. Each element $x_{j}$ in the encoding is a vector of binary values $x_{j}$ = $(b_{1},b_{2},\ldots,b_{L})$, where $L$ is the number of appliances considered in the planning and each value $b_{i}$ indicates if the appliance is on on timestep $j$.
%La representación contempla para cada usuario un vector $X = (x_{0}, x_{1},\ldots,x_{T})$ donde $T$ es la cantidad de minutos en el periodo de tiempo total a planificar. Cada elemento $x_{j}$ es un vector de bits de la forma $x_{j}$ = $(b_{1},b_{2},\ldots,b_{L})$ donde $L$ es la cantidad de dispositivos a planificar y cada $b_{i}$ indica si el dispositivo esta encendido en el minuto $j$.

\begin{figure}[!h]
\setlength{\abovecaptionskip}{0pt}
\setlength{\belowcaptionskip}{2pt}
  \centering
  \includegraphics[width=0.875\textwidth]{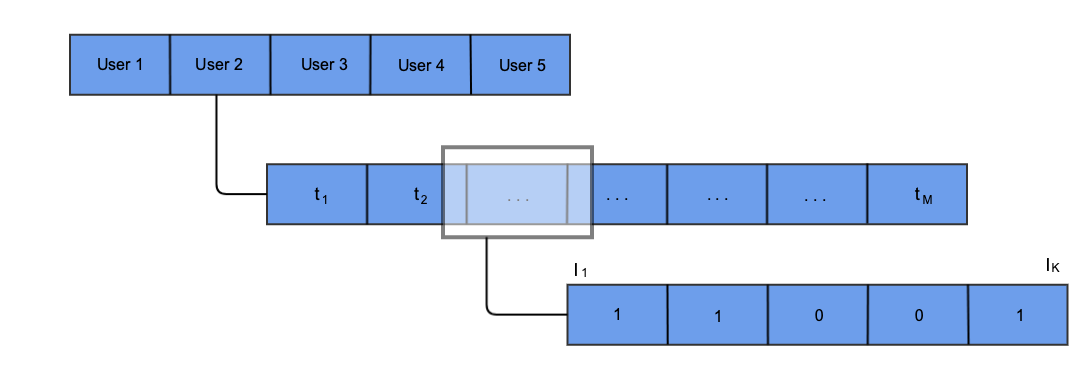}
  \caption{An example of the proposed solution encoding.}
  \label{fig:representacion}
\end{figure}

Fig.~\ref{fig:representacion} presents an example of solution encoding for an instance of the % household appliances planning 
problem considering five appliances. In the example, at timestep (minute) $i$, appliances \#1, \#2, and \#5 are ON, while appliances \#3 and \#4 are OFF.

\textit{Fitness assignment.}
The fitness function of the proposed EA corresponds to an linear aggregation of the 
%two functions defined in Section~\ref{Sec:Prob} (power consumption and user preference), as defined by Eq.~\ref{Eq:fitness}.
power consumption and user 
%preference 
satisfaction
functions:
%(Eq.~\ref{Eq:fitness}).
%\begin{equation}
%\label{Eq:fitness}
$
F = \alpha \times f(X) - \beta \times g(X)
$.
%\end{equation}
Several combinations of weights ($\alpha$, $\beta$) were studied in order to properly weight each objective function and provide a useful search pattern. The main results of the analysis are reported in Section~\ref{Sec:weights}. The combination that allowed computing the best results was ($\alpha=0.65$, $\beta=1$).

\vspace{2mm}
\textit{Initialization.}
The population of tentative solutions is initialized by applying a randomized 
%procedure 
method
that assigns to each appliance a probability $\gamma = 0.6$ 
(value tuned in preliminary experiments) 
to be ON at each time step, following a discrete non-uniform distribution.
%The value of $\gamma$ was tuned in preliminary experiments.
%that confirmed that 
Assigning a slightly larger probability to each appliance to be ON than to be OFF allows starting the evolutionary search for a more diverse set of solutions. 
%The main reason for this decision is that turning off an electric appliance implies reducing the total consumed power (and thus, the cost of the electric bill), while turning on a device implies improving the user satisfaction. 
The value of $\gamma$ was set to provide an equal pressure to both objectives, considering the weights defined in the previous paragraph.
% Esto se debe a que el apagado de dispositivos influye en mayor parte al fitness que el encendido. Apagar un dispositivo eléctrico implica la disminución de la factura eléctrica y de la potencia contratada. Encender un dispositivo favorece únicamente la preferencia del usuario. Por lo tanto en las instancias iniciales los dispositivos eléctricos son encendidos con mayor probabilidad, de manera que el algoritmo explore las demás soluciones. 

\vspace{2mm}
\textit{Selection.}
The standard tournament selection was applied in the proposed EA. Preliminary experiments demonstrated that tournament selection provides an appropriate selection presure to guide the search. After a preliminary configuration analysis, the size of the tournament was set to two individuals, and the best of them is selected. 
%Se utilizó el operador de selección por torneo para otorgar una apropiada presión selectiva a las subpoblaciones. Entre los beneficios que se encuentran de utilizar este tipo de selección esta la velocidad de aplicación (dado que no es necesario evaluar ni comparar la totalidad de la población) y la capacidad de prevenir, en cierto grado, la convergencia prematura.

\vspace{2mm}
\textit{Evolutionary operators.}
Ad-hoc evolutionary operators were conceived to provide efficacy and diversity to the search, working with the proposed solution encoding.
The proposed evolutionary operators are:
\begin{itemize}
    \item \textit{Recombination}. An ad-hoc version of the Single Point Crossover 
    %(SPX) 
    operator was conceived to recombine solutions. A cutting point is selected for each user and a new planning is created for each user, using information from the first parent (before the cutting point) and from the second parent (after the cutting point).
%Para recombinar los individuos se utilizó el operador de recombinación de un punto (SPX) para cada usuario. La figura \ref{fig:recombinacion-ejemplo} muestra un ejemplo de recombinación para tres usuarios con sus respectivos dispositivos eléctricos.
%Dado dos elementos de la población (individuos padres), para cada usuario, se selecciona de forma aleatoria el intervalo de tiempo donde realizar la recombinación (punto de corte). A partir del intervalo seleccionado, el individuo resultante forma una nueva planificación de uso de dispositivos eléctricos utilizando la información antes de el punto de corte de uno de los individuos padres y la información luego del punto de corte del otro individuo padre.
Fig.~\ref{fig:recombinacion-ejemplo} presents an example of the application of the proposed recombination operator between two solution for a problem instance with three users and six appliances for each user.

\begin{figure}[!h]
\setlength{\abovecaptionskip}{0pt} \makebox[\textwidth][c]{\includegraphics[width=\textwidth]{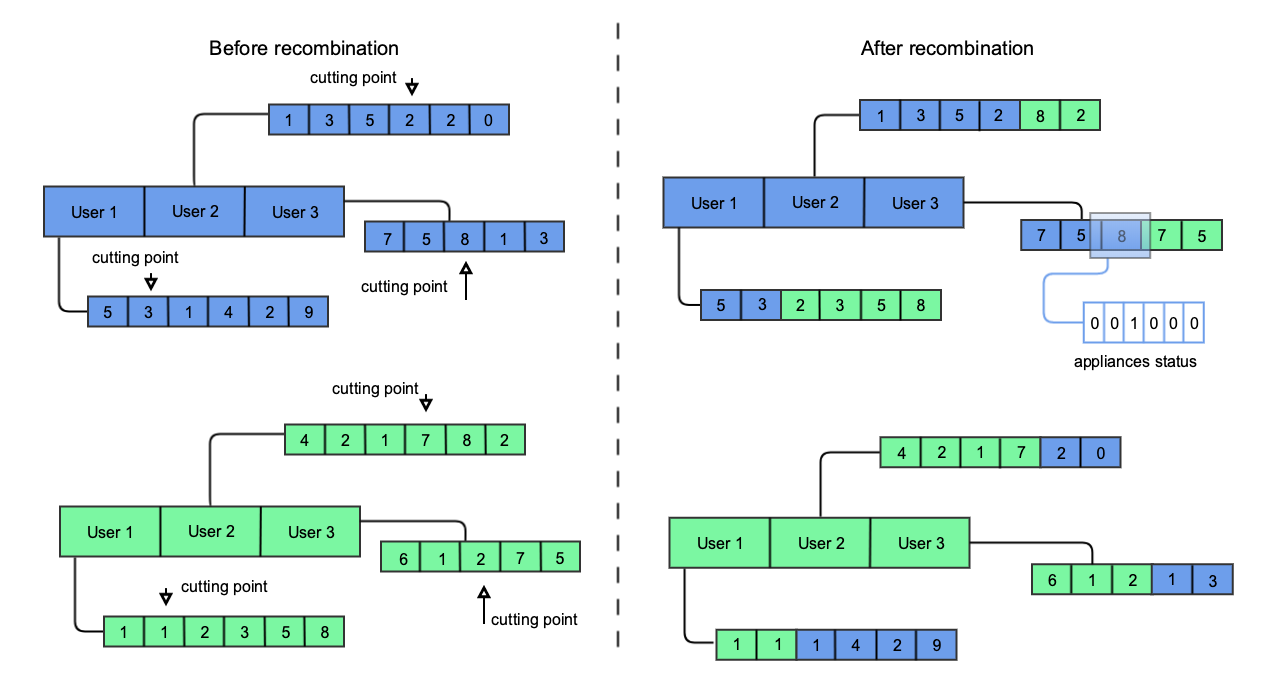}}%
  \caption{An example of the proposed recombination operator.}
  \label{fig:recombinacion-ejemplo}
\end{figure}

\textit{Mutation.} The mutation operator 
%is based on modifying 
modifies
the current state of an appliance. First, a specific time interval is randomly selected for every user, according to a uniform distribution. An appliance is then randomly selected (applying a uniform distribution) from all belonging to that user, and its state is changed (on/off or viceversa). Fig.~\ref{fig:mutacion-ejemplo} presents an example of the mutation operator.

%El operador de mutación cambia el estado actual de un dispositivo. Si el dispositivo se encuentra encendido, el operador de mutación lo apaga. La selección del intervalo de tiempo donde realizar la mutación como el dispositivo eléctrico el cual mutar se realiza de forma aleatoria. Experimentos iniciales mostraron que cambiar mayor cantidad de dispositivos reducía la diversidad en la población. La figura \ref{fig:mutacion-ejemplo} muestra un ejemplo de mutación para tres usuarios con sus respectivos dispositivos eléctricos.
 
\begin{figure}[!ht]
\setlength{\abovecaptionskip}{0pt}
  \makebox[\textwidth][c]{\centering\includegraphics[width=\textwidth]{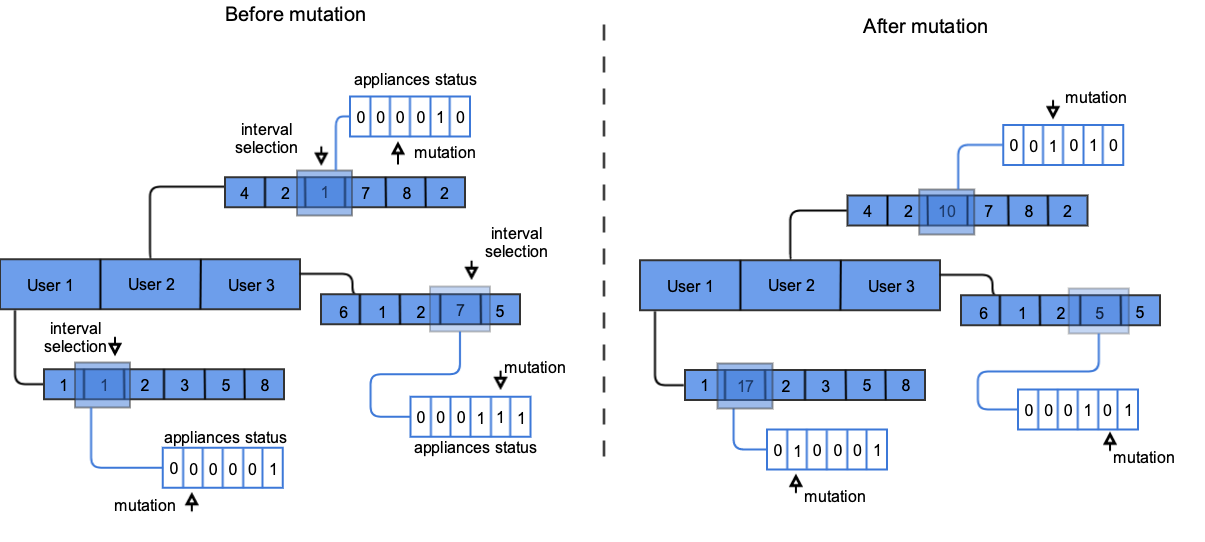}}%
  \caption{An example of the proposed mutation operator.}
  \label{fig:mutacion-ejemplo}
\end{figure}

\end{itemize}

\subsection{Development and execution platform}
The 
%proposed 
evolutionary approach was implemented using the ECJ library,
%ECJ is 
a Java-based evolutionary computation 
%research 
system developed at 
%ECLab Evolutionary Computation Laboratory, 
George Mason University (\url{cs.gmu.edu/~eclab/projects/ecj}). ECJ includes easily modifiable classes for solving optimization problems.
The experimental evaluation was performed on a Dell Power Edge server, Quad-core Xeon E5430 processor at 2.66GHz, 8 GB RAM, 
%Gigabit Ethernet, 
from Cluster FING,
% the high-performance computing facility (
Universidad de la Rep\'ublica, Uruguay 
%http://www.fing.edu.uy/cluster)~
\cite{Nesmachnow2010}.
%from Cluster FING, Universidad de la Rep\'ublica, Uruguay~\cite{Nesmachnow2010}}.

%\blue{??? The experimental evaluations were carried out using ClusterUY (site: https://cluster.uy) that counts with 1120 kernels Intel Xeon-Gold 6138, 3,5 TB RAM memory and 28 GPU Nvidia Tesla P100 with 12Gb of memory interconnected by a high-speed 10 Gbps Ethernet network. ???}

\section{Experimental analysis}
\label{Sec:Exp}

This section presents the experimental analysis of the proposed EA for household energy consumption planning.

\subsection{Problem instances}

A set of problem instances was build using real data. 

%\textit{Data selection.}
Raw energy consumption data for appliances were obtained from the REDD dataset~\cite{Kolter2011}, sampled every three seconds for a period of one month. %In order to select the appliances to consider in the problem instances, the following procedure was applied: 
%LARGO
\if 0
The following procedure was applied to select the appliances to include in the problem instances:
i) the categorization of household appliances according to their operating profiles and purposes~\cite{Soares2013} was taken into account; ii) the average and maximum time of use of each appliance was computed to determine those appliances that mostly contribute to the overall energy consumption; and iii) several other parameters were considered, including the number of activations and energy consumption measurements, the number of houses where each appliance is present, the frequency of activation, etc. Six appliances were selected: dishwasher, microwave, dryer, air conditioning, oven, and refrigerator.
\fi
% CORTO
Appliances included in the problem instances were selected considering:
i) the categorization of household appliances according to their operating profiles and purposes~\cite{Soares2013};
%was taken into account; 
ii) the average and maximum time of use of each appliance; 
%was computed to determine those appliances that mostly contribute to the overall energy consumption; 
%(selecting those that most contribute to the overall energy consumption);
and iii) several other parameters, 
%were considered, 
including the number of activations and energy consumption measurements, the number of houses where each appliance is present, the frequency of activation, etc. Six appliances were selected: dishwasher, microwave, dryer, air conditioning, oven, and refrigerator.
The planning period is one day.

%For defining the user preference function, 
Energy consumption data 
%available 
in the REDD dataset was averaged over intervals of 15 minutes, and the resulting values were analyzed %in order 
to define the preferences of using each %household 
appliance for each user.
The overall energy consumption data were studied for each day.
%, separating weekdays from weekend. 
One representative weekday and one representative weekend day was determined for each appliance, in order to define the daily power consumption in each case.
%
%Finally, 
%The energy price correspond to real tariffs
Real energy prices
from the National Electricity Company (UTE) in Uruguay (\url{https://portal.ute.com.uy}), considering an average residential plan
%, with a contracted power of 3.7 kWh.
(contracted power of 3.7 kWh) were used.
%Para las instancia de edificio la energía eléctrica contratada se calcula como la energía eléctrica contratada por un usuario por la cantidad de usuarios en el edificio.

%\textit{Instances generation.}
Six problem instances were generated
(Table~\ref{Tab:instances})
%: two \textit{small instances}, two \textit{medium instances}, and two \textit{large instances}, 
accounting for different number of users, 
appliances, and consumption patterns (weekday/weekend). 

\begin{table}[!h]
\setlength{\tabcolsep}{6pt}
\renewcommand{\arraystretch}{0.85}
\setlength{\abovecaptionskip}{0pt}
\setlength{\belowcaptionskip}{0pt}
\centering
\small
\caption{Proposed problem instances}
\label{Tab:instances}
\begin{tabular}{rlrrr}
\toprule

\multicolumn{1}{c}{{\#}} & \multicolumn{1}{c}{\textit{name}} & \multicolumn{1}{c}{\textit{users}} & 
\multicolumn{1}{c}{\textit{appliances in the instance}} & \multicolumn{1}{c}{\textit{consumption pattern}} \\
\midrule
1 & small.1 (s1) & 2 & (2,3) & weekday \\
2 & small.2 (s2) & 2 & (2,3) & weekend \\
\midrule
3 & medium.1 (m1) & 4 & (4,4,3,2) & weekday \\
4 & medium.2 (m2) & 4 & (4,4,3,2) & weekend \\
\midrule
5 & large.1 (l1) & 6 & (5,5,4,4,3,2) & weekday \\
6 & large.2 (l2) & 6 & (5,5,4,4,3,2) & weekend \\
\bottomrule
\end{tabular}
\end{table}

\subsection{Baseline algorithms for results comparison}

Two baseline strategies were implemented for evaluating the results of the proposed EA for household appliances planning: a greedy algorithm and a Business-as-Usual (BaU) planning strategy, which are described next.

\vspace{1mm}
\textit{Greedy planning strategy.}
Greedy algorithms 
%are iterative methods to build 
iteratively build
solutions based on a
%heuristic focused on 
taking 
optimal local decisions in each step.
%, with the aim of approximating the optimal general solution of the problem.
%El algoritmo ávido implementado buscó maximizar una función lineal que tuviese en cuenta la preferencia del usuario y el precio de la energía eléctrica en cada instante de tiempo.
%Para cada usuario y para cada dispositivo eléctrico, se buscó obtener el intervalo de tiempo que maximiza la ecuación \ref{eq:monoobjetivo}. Específicamente, el algoritmo buscó encontrar un intervalo de preferencia  en el que la energía consumida no se vea superada por la energía eléctrica contratada por el usuario. Para ese intervalo se obtiene el valor obtenido de aplicar la ecuación \ref{eq:monoobjetivo} (PrefCosto). 
%
A 
%simple 
greedy algorithm 
from the literature~\cite{Orsi2017} 
was 
%designed and implemented 
adapted
for baseline comparison. 
The proposed strategy 
%works by searching 
searches
%appropriate 
the best
time intervals to switch on each appliance $d_k$, according to the user 
%preferences 
satisfaction
and cost
%The schema of the greedy algorithm is presented on Algorithm~\ref{Alg:greedy}.
%(Algorithm~\ref{Alg:greedy}). The best interval for using appliance $d$ is searched
, considering the linear aggregation fitness function using $\alpha=0.65$, $\beta=1$ (Algorithm~\ref{Alg:greedy}).

\begin{algorithm}[!h]
\setstretch{0.95}
\footnotesize
\caption{Greedy algorithm for household appliances planning}
\label{Alg:greedy}

\begin{algorithmic}

\Procedure{IntervalMaxPrefCost}{initMin,u$_\text{i}$,d,X}
\State prefCost $\leftarrow$ 0; duration $\leftarrow$ 0
%\State duration $\leftarrow$ 0
\For{(m=initMin; m$<t_M$; m++)}
    \If{duration $< D(d,u_i)$}
        \If{$\sum_{k=1}^{K} x^i_{km} \times P(d_{k}) + P(d) < E(u_i)$}
            \State prefCost $\leftarrow$ prefCost + $\alpha$ $\times $ $UP(u,d,m) - \beta$ $\times$ $C(m)$ 
            \State duration $\leftarrow$ duration + ($t_{m+1} - t_{m}$)
        \Else
            \State prefCost $\leftarrow$ 0
            \State duration $\leftarrow$ 0
        \EndIf
    \Else
        \State \Return [m, prefCost] \Comment{\blue{interval found}}
    \EndIf
\EndFor
\State \Return [m, prefCost] \Comment{\blue{no interval found}}
\EndProcedure
\\
\State X $\leftarrow \overrightarrow{\vec{0}}$
\For{($i=1$; $i\le$N; $i$++)} \Comment{\blue{for each user}}
%    \State m $\leftarrow$ $t_1$
    \For{(k=1;i$\le$K;k++)} \Comment{\blue{for each appliance}}
        \State prefCost $\leftarrow$ 0; bestPrefCost $\leftarrow$ -1; bestmin $\leftarrow$ 0 \Comment{\blue{search best interval}}
%        \State bestPrefCost $\leftarrow$ -1
%        \State bestmin $\leftarrow$ 0
%        \While{prefCost $\neq$ 0 and prefCost $>$ bestPrefCost}
        \For{(m=$t_1$; m$<t_M-D(d_k,u_i)$; m++)}
            \State [min, prefCost] = IntervalMaxPrefCost(m, $d_k$, $u_i$, X)
            \If{prefCost $>$ bestPrefCost}
                \State bestPrefCost $\leftarrow$ prefCost
                \State bestmin $\leftarrow$ min

            \EndIf
%        \EndWhile
        \EndFor
        \For{(m=bestmin $- D(d_k,u_i)$; m $\le$ bestmin; m++)}
            \State $x^i_{km} \leftarrow$ 1 \Comment{\blue{set appliance ON}}
        \EndFor
    \EndFor
\EndFor
\end{algorithmic}
\end{algorithm}

%\vspace{1mm}
\textit{BaU planning strategy}. 
The BaU strategy proposes assigning ON times to each appliance without planning. These plannings have good user preference values, but suboptimal cost values.
%La estrategia \textit{Business as Usual} (BaU) propone una planificación exactamente igual al consumo actual del usuario. Para todas las instancias evaluadas se realizo también la planificación BaU y se estudio el resultado de aplicar las funciones definidas en (\red{refereciar-formulacion-matematica}). Es relevante comparar el desempeño del MOEA frente a la solución de BaU de forma de tener un valor comparativo frente a la planificación actual del usuario.

\subsection{Linear aggregation fitness function}
\label{Sec:weights}

The analysis considered 
%{four candidate values} for $\alpha$ (0.3, 0.65, 0.75, and 1.0) and {three candidate values} for $\beta$ (0.3, 0.5, and 1.0). 
%the following candidate values: $\alpha \in \{ 0.3, 0.65, 0.75, 1.0\}$ and $\beta \in \{0.3, 0.5, 1.0\}$. 
the candidate values $\alpha \in \{ 0.3, 0.65, 0.75, 1.0\}$ and $\beta \in \{0.3, 0.5, 1.0\}$. 
%
%The first stage involved a non-formalized sensitivity analysis to select those combinations that allowed computing the best (i.e., smallest) fitness values for a randomly generated medium-size instance. After that, six configurations were selected: (0.3,0.5); (0.4,0.5); (0.5,0.5); (0.5,0.7); (0.5,1.0); and (0.65,1.0), to perform an exhaustive analysis by executing the proposed EA over three medium-size instances of the problem (one house with two, four, and six devices).
%The analysis was performed by executing the proposed EA over
The 
%proposed 
EA was executed
over three medium-size instances of the problem 
%(one house with 
(two, four, and six devices).
Table~\ref{Tab:resultadoslineal} reports the mean and interquartile range (IQR) of the best fitness value computed in 30 independent executions of the proposed EA for the three instances solved, using 
%each one of the six candidate configurations for the linear aggregation fitness function.
the studied configurations.
%
%The best results (smallest absolute value) were obtained using the configuration $\alpha=0.65$ and $\beta=1$. This configuration was used in both parameter setting and validation experiments. 

\begin{table}[!h]
\setlength{\tabcolsep}{8pt}
\setlength{\abovecaptionskip}{3pt}
\setlength{\belowcaptionskip}{0pt}
\renewcommand{\arraystretch}{0.85}
\centering
\small
\caption{Best fitness values computed using different values of $(\alpha,\beta)$}
\label{Tab:resultadoslineal}
\begin{tabular}{rrrrrrrrrr}
\toprule
\multirow{2}{*}{$\alpha$} & 
\multirow{2}{*}{$\beta$} & \multicolumn{2}{c}{instance \#1} & &  \multicolumn{2}{c}{instance \#2} & & 
\multicolumn{2}{c}{instance \#3} \\
\cline{3-4}\cline{6-7}\cline{9-10}\\[-7pt]
& & median & IQR & & median & IQR & & 
median & IQR \\
\midrule
0.3 & 0.3 & 15.81 & 1.09 & & 179.83 & 0.55 & & 132.35 & 6.68\\
0.3 & 0.5 & 11.02 & 1.54 & & 167.94 & 2.03 & & 110.10 & 4.54 \\
0.3 & 1.0 & 4.27 & 6.39 & & 128.55 & 28.66 & & 54.82 & 13.56 \\
0.5 & 0.3 & 34.06 & 0.81 & & 311.32 & 28.66 & & 253.95 & 5.11 \\
0.5 & 0.5 & 26.75 & 1.06 & & 298.80 & 23.71 & & 224.47 & 12.21 \\
0.5 & 1.0 & 13.53 & 10.28 & & 269.83 & 26.03 & & 175.11 & 40.39 \\
0.65 & 0.3 & 47.28 & 2.08 & & 409.91 & 1.76 & & 342.66 & 6.57 \\ 
0.65 & 0.5 & 39.87 & 0.83 & & 398.45 & 1.34 & & 316.32 & 8.40 \\
0.65 & 1.0 & 25.48 & 6.12 & & 369.21 & 3.91 & & 257.55 & 39.03 \\
0.75 & 0.3 & 54.85 & 2.79 & & 475.64 & 0.39 & & 405.86 & 10.03 \\
0.75 & 0.5 & 48.52 & 3.20 & & 462.74 & 35.73 & & 373.95 & 4.11 \\
0.75 & 1.0 & 30.18 & 3.43 & & 433.82 & 3.23 & & 304.31 & 21.64 \\
1.0 & 0.3 & 81.21 & 4.41 & & 640.62 & 0.59 & & 548.20 & 5.64 \\
1.0 & 0.5 & 71.46 & 4.55 & & 628.42 & 2.76 & & 524.55 & 9.10 \\
1.0 & 1.0 & 52.05 & 2.89 & & 597.33 & 0.24 & & 454.48 & 20.50 \\
\bottomrule
\end{tabular}
\end{table}

Fig.~\ref{Fig:trade-off} presents a 
%graphical comparison of the 
trade-off analysis  of solutions computed using different values of 
%parameters 
($\alpha$,$\beta$) 
%in the linear aggregation fitness function 
for instance \#3 (results are representative of those obtained for other instances). 
%Results demonstrate that 
The combination (0.65,1.0) allows computing the best trade-off solutions regarding user satisfaction and total energy/price.

\if 0
\begin{figure}[!h]
  \centering
\begin{tikzpicture}
	\begin{axis}[%
	xlabel = user satisfaction $f(X)$,
	ylabel = total energy/price $g(X)$,
	xmin=345,
    xmax=480,
    ymin=48,
    ymax=110,
	scatter/classes={%
		a={mark=o,blue},%
		b={mark=square,draw=green}}]
	\addplot[scatter,only marks,%
		scatter src=explicit symbolic,%
		nodes near coords*={\Label},
		visualization depends on={value \thisrow{label} \as \Label},
		visualization depends on={value \thisrow{anchor}\as\myanchor},
   every node near coord/.append style={anchor=\myanchor,scale=0.75},
%   every node near coord/.append style={font=\small}
] table[meta=class] {tradeoff_inactive.dat};
	\end{axis}
    \draw [red, ->] (1,4.55) -- (4,2);
    \node[red,scale=0.85] at (2.8,4) {best solutions};
\end{tikzpicture}
  \caption{Trade-off analysis of solutions computed using different values of parameters ($\alpha$,$\beta$) in the linear aggregation fitness function (\blue{$\circ$}--non-dominated solutions, \green{$\square$}--dominated solutions) for instance \#3 (representative case).}
  \label{Fig:trade-off}
\end{figure}
\fi

\begin{figure}[!h]
\setlength{\abovecaptionskip}{0pt}
\setlength{\belowcaptionskip}{0pt}
\centering
\begin{tikzpicture}
\pgfplotsset{every axis/.append style={
%                    axis x line=middle,    % put the x axis in the middle
%                    axis y line=middle,    % put the y axis in the middle
%                    axis line style={->}, % arrows on the axis
                    label style={font=\footnotesize},
                    tick label style={font=\footnotesize}  
                    }
}
	\begin{axis}[%
	legend style={at={(0.35,0.95)},anchor=north},
	legend cell align={left},
	xlabel = \footnotesize user satisfaction $f(X)$,
	ylabel = \footnotesize total energy/price $g(X)$,
	xmin=345,
    xmax=480,
    ymin=48,
    ymax=110,
	scatter/classes={%
		a={mark=o,blue},%
		b={mark=square,draw=green}}]
	\addplot[scatter,only marks,%
		scatter src=explicit symbolic,%
		nodes near coords*={\Label},
		visualization depends on={value \thisrow{label} \as \Label},
		visualization depends on={value \thisrow{anchor}\as\myanchor},
   every node near coord/.append style={anchor=\myanchor,scale=0.75},
%   every node near coord/.append style={font=\small}
] table[meta=class] {tradeoff_active.dat};
	\legend{\scriptsize ~non-dominated solutions, \scriptsize ~dominated solutions}
	\end{axis}
    \draw [red, ->,-{Latex[length=3mm]}] (1.0,4.3) -- (4,2);
    \node[red,scale=0.85] at (3.2,3.5) {best solutions};
\end{tikzpicture}
\if 0
\caption{Trade-off analysis of solutions computed using different values of %parameters 
  ($\alpha$,$\beta$) in the linear aggregation fitness function 
  %(\blue{$\circ$}--non-dominated solutions, \green{$\square$}--dominated solutions) 
  for instance \#3 (representative case).}
\fi
\caption{Trade-off analysis of solutions computed using different values of ($\alpha$,$\beta$)}
\label{Fig:trade-off}
\end{figure}

\subsection{Parametric configuration analysis}

%Due to the stochastic nature of EAs, its  
EA
parameters must be adjusted 
%before the experimental analysis, 
to determine the configuration that allows computing the best results. The analysis was performed over three problem instances, different from those used in the 
%experimental 
evaluation to avoid bias. After an initial evaluation, the population size was fixed at 150 individuals.

Three relevant parameters of the proposed EA were studied: number of generations used as stopping criterion ($G$), recombination probability $p_{C}$ and mutation probability $p_{M}$. Candidate values for each parameter were: $p_{C} \in \{0.1, 0.25, 0.5\}$; $p_{M} \in \{0.1, 0.05, 0.01\}$; and $G \in \{2500, 5000, 10000\}$.
All combinations of parameter values were studied by performing 50 independent executions of the proposed EA for the three problem instances considered in the analysis. The metric considered in the analysis was the linear aggregation fitness function defined in the previous subsection.

The methodology for selecting the best configuration included: i) the Shapiro-Wilk statistical test was applied to check normality, taken as a null hypothesis that the results followed a normal distribution; as p-values less than 0.05 were obtained, the null hypothesis was discarded and 
%the fitness results were considered as corresponding to a non-normal distribution;  
it was assumed that the fitness results follow a non-normal distribution;
ii) the Friedman's rank test was applied, taken as a null hypothesis that the fitness distributions for the different configurations were not different, as p-values less than 0.05 were obtained, the null hypothesis was discarded and the 
%fitness 
results significantly differ from each other.
%A tournament was held between the best fitness values ​​and the best standard deviation, selecting as the best configuration the one that achieved the best fitness value with the least standard variation.

Table~\ref{Tab:DetalleConfParametricaAE} reports the fitness values computed in the parameter setting experiments for a representative problem instance.
%Fig.~\ref{fig:resultadoConfParametricaAE} graphically summarizes the main results of the parameter setting experiments. 
%
Overall, the best results 
%in the parameter setting experiments
(i.e., largest fitness median and also lower IQR) were obtained using configuration \#11 (values $G = 10000$, $p_{C} = 0.1$, and $p_{M} = 0.1$). Henceforth, these values were used in the validation experiments of the proposed EA.

\begin{table*}[!h]
\setlength{\tabcolsep}{6pt}
\setlength{\abovecaptionskip}{3pt}
\renewcommand{\arraystretch}{0.9}
\centering
\small
\caption{Parameter setting results for the proposed EA}
\label{Tab:DetalleConfParametricaAE}
\begin{tabular}{rrrrrr}
\toprule
\multicolumn{1}{c}{\textit{configuration}} & \multicolumn{2}{c}{\textit{fitness}} & \multicolumn{1}{c}{\textit{configuration}} & \multicolumn{2}{c}{\textit{fitness}} \\
\cline{2-3} \cline{5-6} \\[-9pt]
\multicolumn{1}{c}{($G$, $p_C$, $p_M$)} & \multicolumn{1}{c}{\textit{median}} & \multicolumn{1}{c}{\textit{IQR}} &
\multicolumn{1}{c}{($G$, $p_C$, $p_M$)} & \multicolumn{1}{c}{\textit{median}} & \multicolumn{1}{c}{\textit{IQR}} \\
\midrule
(2500,  0.1, 0.01) &  7.88 & 10.45 & (10000, 0.5,  0.1) & 40.27 & 7.61  \\
(5000,  0.1, 0.01) & 38.74 & 11.71 & (2500,  0.25, 0.1) & 40.73 & 7.01  \\
(10000, 0.1, 0.01) & 40.04 & 6.98  & (5000,  0.25, 0.1) & 41.64 & 6.76 \\
(2500,  0.5, 0.01) &  9.26 & 11.17 & (10000, 0.25, 0.1) & 40.72 & 8.67  \\
(5000,  0.5, 0.01) & 39.22 & 9.37 &  (2500,  0.1, 0.05) & 39.60 & 6.97 \\
(10000, 0.5, 0.01) & 39.28 & 9.88 &  (5000,  0.1, 0.05) & 40.22 & 9.23 \\
(2500,  0.25,0.01) &  9.06 & 9.22 &  (10000, 0.1, 0.05) & 40.58 & 7.95  \\
(5000,  0.25,0.01) & 39.39 & 10.18 & (2500,  0.5, 0.05) & 38.66 & 7.39 \\
(10000, 0.25,0.01) & 38.69 & 11.76 & (5000,  0.5, 0.05) & 39.59 & 10.07 \\
(2500,  0.1,  0.1) & 41.77 & 8.88 & (10000,  0.5, 0.05) & 40.79 & 11.62 \\
(5000,  0.1,  0.1) & 41.30 & 8.46 & (2500,  0.25, 0.05) & 38.17 & 10.32 \\
\textbf{(10000, 0.1,  0.1)} & \textbf{42.57} & \textbf{6.46} & (5000,  0.25, 0.05) & 41.67 &  8.38 \\
(2500,  0.5,  0.1) & 40.62 & 7.22 & (10000,  0.25, 0.05) & 41.00 & 8.44 \\
(5000,  0.5,  0.1) & 41.28 & 8.65 \\
\bottomrule
\end{tabular}
\end{table*}

\subsection{Experimental results}

Table \ref{Tab:ResultadosAE} reports the median of the best fitness 
%values 
computed by the 
%proposed 
EA and the comparison with the reference algorithms (greedy and BaU). The relative improvement on fitness values ($\Delta_f$) and on each objective function ($\Delta_\text{cost}$, $\Delta_\text{pref}$) over each reference algorithm refA is computed as $\Delta = (f$(EA)$-f$(refA))/$f$(refA).

\if 0
\begin{table}[!h]
\setlength{\tabcolsep}{6pt}
\setlength{\abovecaptionskip}{3pt}
\renewcommand{\arraystretch}{0.9}
\centering
\small
\begin{tabular}{lrrrrr}
\toprule 
\multicolumn{1}{c}{\textit{instance}} & $f$(EA) & $f$(greedy) & $\Delta$(greedy) & $f$(BaU) & $\Delta$(BaU) \\
\midrule
\multicolumn{6}{c}{\textit{weekday}}\\
\midrule
\multicolumn{1}{c}{small.1} & 14.07 & 10.97 & 28\% \\
\multicolumn{1}{c}{medium.1} & 340.02 & 239.41 & 42\% \\
\multicolumn{1}{c}{large.1} & 348.48 & 347.98 & 0.14\% \\
\midrule
\multicolumn{6}{c}{\textit{weekend}}\\
\midrule
\multicolumn{1}{c}{small.2} & 323.67 & 252.08 & 28\% \\
\multicolumn{1}{c}{medium.2} & 253.39 & 197.18 & 28\% \\
\multicolumn{1}{c}{large.2} & 369.81 & 351.90 & 5\% \\
\bottomrule
\end{tabular}
\caption{Experimental results: fitness values and improvements of the proposed EA over the baseline greedy algorithm and the BaU strategy.}
\label{Tab:ResultadosAE}
\end{table}
\fi

\begin{table}[!h]
\setlength{\tabcolsep}{4pt}
\setlength{\abovecaptionskip}{2.5pt}
\renewcommand{\arraystretch}{0.9}
\centering
\small
\caption{Experimental results: fitness values and improvements of the proposed EA over the baseline greedy algorithm and the BaU strategy.}
\label{Tab:ResultadosAE}
\begin{tabular}{crrrrrrrrrr}
\toprule 
\multirow{2}{*}{{\textit{instance}}} & \multirow{2}{*}{$f$(EA)} & \multicolumn{4}{c}{\textit{greedy}} & & \multicolumn{4}{c}{\textit{BaU}}\\
\cline{3-6}\cline{8-11}\\[-8pt]
& & \multicolumn{1}{c}{$f$} & \multicolumn{1}{c}{$\Delta_f$} & \multicolumn{1}{c}{$\Delta_\text{cost}$} & \multicolumn{1}{c}{$\Delta_\text{pref}$} & & 
\multicolumn{1}{c}{$f$} & \multicolumn{1}{c}{$\Delta_f$} & \multicolumn{1}{c}{$\Delta_\text{cost}$} &  \multicolumn{1}{c}{$\Delta_\text{pref}$} \\
\midrule
\multicolumn{10}{c}{\textit{weekday}}\\
\midrule
\multicolumn{1}{c}{small.1} & 14.1 & 11.0 & 28.3\% & 22.6\% & 56.2\% & &
-7.6 & 284.6\% & 81.1\% & 16.0\%\\
\multicolumn{1}{c}{medium.1} & 340.0 & 239.4 & 42.0\% & 27.2\% & 65.3\% & & 69.5 & 388.9\% & 77.9\% & 19.4\%\\
\multicolumn{1}{c}{large.1} & 407.8 & 347.9 & 17.2\% & 20.8\% & 47.3\% & & -187.1 & 317.0\% & 70.6\% & 22.0\%\\
\midrule
\multicolumn{10}{c}{\textit{weekend}}\\
\midrule
\multicolumn{1}{c}{small.2} & 323.7 & 252.1 & 28.4\% & 25.1\% & 44.9\% & & 67.4 & 383.1\% & 76.6\% & 25.8\% \\
\multicolumn{1}{c}{medium.2} & 253.4 & 197.2 & 28.5\% & 29.0\% & 48.1\% & & 153.8 & 64.7\% & 60.6\% & 20.8\% \\
\multicolumn{1}{c}{large.2} & 369.8 & 351.9 & 5.1\% & 19.7\% & 37.4\% & & -299.9 & 224.2\% & 72.2\% & 31.6\% \\
\bottomrule
\end{tabular}
\end{table}

Results in Table~\ref{Tab:ResultadosAE} indicate that the proposed EA is able to improve significantly over the greedy algorithm regarding the fitness values. 

Considering the baseline results computed by the proposed greedy algorithm, improvements of up to 42.0\% were obtained in instance medium.1. Results also suggest that consumption patterns during the weekend are harder to plan for the EA, as the improvements over the greedy algorithm reduced to 5.1\% in instance large.2. This can be explained due to the interactive utilization of household appliances in weekends, when people are at home a significantly larger periods than in weekdays. Regarding the improvements on user satisfaction and cost, the plannings computed by the proposed EA allow reducing more than 20\% the electric bill, and preferences improve more than 40\% in all the studied scenarios. 

The EA computed significantly cheaper plannings than those of BaU, which 
systematically failed to provide good cost values, indicating that users do not take the correct decisions to turn on home appliances in this regard, and they can benefit of having an automated planning offered by a recommendation system. In addition, preferences on the solutions computed by the EA were 16--31\% better than BaU. The obtained improvements over a BaU strategy are consistent with results reported in previous works for a reduced subset of home appliances (air conditioner and water heater)~\cite{Orsi2017}.

The obtained results suggest that the proposed evolutionary approach is accurate for computing household energy consumption plannings accounting for both energy costs and user satisfaction at the same time.
The proposed approach is a first step towards designing an intelligent recommendation system for end-users.

\if 0
%\red{Analysis of plannings}
% Fig.~\ref{fig:fit_inst_1} and~\ref{fig:fit_inst_2} show ..

\begin{figure}
\centering
%\includegraphics[scale=0.4]{p1}
\caption{.... asd.}
\label{fig:fit_inst_1}
\end{figure}
\fi

\section{Conclusions and future work}
\label{Sec:Conc}

This article presented an evolutionary approach to address the problem of household energy planning, as a first approach to develop an automated recommendation system for end-users. This is a relevant problem for both energy companies and citizens under the novel smart city paradigm.

A mathematical formulation for the problem was proposed, considering the optimization of user preferences and energy consumed/cost. A specific EA was proposed to solve the problem, simultaneously optimizing both criteria using a linear aggregation multiobjective function and ad-hoc evolutionary operators.

A set of six realistic problem instances built using real data were considered in the experimental evaluation of the proposed EA. The analysis compared the EA results with two baseline planning methods (greedy and business-as-usual).

The experimental results showed that the proposed EA is able to compute accurate plannings, accounting for significant improvements on the problem objectives. Regarding the baseline greedy algorithm, improvements of up to 42.0\% were obtained in the proposed multiobjective function, accounting for an average reduction of more than 20\% in the energy consumption (and thus, on the electric bill) and preferences improved more than 40\% in all the studied scenarios. Regarding the BaU strategy, the EA computed significantly cheaper plannings and user preferences improved up to 31\%, in line with previous results from our research group.

The obtained results suggest that the proposed evolutionary approach is accurate for computing household energy consumption plannings accounting for both energy costs and user satisfaction at the same time. Overall, the proposed algorithm showed to be effective for addressing the considered optimization problem. The analysis demonstrated that users can significantly benefit of having an automated planning offered by a recommendation system.

The main lines for future work are related to study explicit multiobjective algorithms to solve the problem, in order to compute several trade-off solutions at the same time. The problem formulation can be extended to include the noisy nature of user preferences in order to define an uncertainty optimization problem. In this regard, robust evolutionary approaches should be studied to solve this problem variant. Finally, new real problem instances can be generated, especially using data from the National Electricity Administration (UTE), in Uruguay.

\bibliographystyle{splncs03}
\bibliography{ref.bib}

\end{document}